\documentclass{article}

\usepackage{arxiv}
\usepackage{graphicx}
\usepackage{newtxtext,newtxmath}
\usepackage[utf8]{inputenc} 
\usepackage[T1]{fontenc}    
\usepackage{hyperref}       
\usepackage{url}            
\usepackage{booktabs}       
\usepackage{amsfonts}       
\usepackage{nicefrac}       
\usepackage{microtype}      
\usepackage{lipsum}

\title{Hierarchical Semantic Correspondence Learning for Post-Discharge Patient Mortality Prediction}

\author{
  Shaika Chowdhury \\
  University of Illinois at Chicago\\
  \texttt{schowd21@uic.edu} \\
   \And
 Chenwei Zhang \\
  Amazon\\
  \texttt{cwzhang@amazon.com} \\
  \And
   Philip S.Yu \\
  University of Illinois at Chicago\\
  \texttt{psyu@uic.edu} \\
   \And
 Yuan Luo \\
  Northwestern University\\
  \texttt{yuan.luo@northwestern.edu} \\ 
}

\begin{document}
\maketitle

\begin{abstract}
Predicting patient mortality is an important and challenging problem in the healthcare domain, especially for intensive care unit (ICU) patients. Electronic health notes serve as a rich source for learning patient representations, that can facilitate effective risk assessment. However, a large portion of clinical notes are unstructured and also contain domain specific terminologies, from which we need to extract structured information. In this paper, we introduce an embedding framework to learn semantically-plausible distributed representations of clinical notes that exploits the \textit{semantic correspondence} between the unstructured texts and their corresponding structured knowledge, known as \textit{semantic frame}, in a hierarchical fashion. Our approach integrates text modeling and semantic correspondence learning into a single model that comprises 1) an unstructured embedding module that makes use of self-similarity matrix representations in order to inject structural regularities of different segments inherent in clinical texts to promote local coherence, 2) a structured embedding module to embed the semantic frames (e.g., UMLS semantic types) with deep ConvNet and 3) a hierarchical semantic correspondence module that embeds by enhancing the interactions between text-semantic frame embedding pairs at multiple levels (i.e., words, sentence, note). Evaluations on multiple embedding benchmarks on post-discharge intensive care patient mortality prediction tasks demonstrate its effectiveness compared to approaches that do not exploit the semantic interactions between structured and unstructured information present in clinical notes.
\end{abstract}


\section{Introduction}
 Identifying the status of high risk patients (i.e., ICU) after discharge from hospital is a critical step in the overall improvement of healthcare systems and better decision making by physicians. It can help healthcare providers follow up the potential causes of death after hospital discharge and expedite treatment \cite{luo2016interpretable}. The rapid growth of digitized clinical data (e.g., electronic health records) offers an abundance of information on patient trajectory. Most recent works on patient mortality prediction \cite{sushil2018patient,grnarova2016neural} have relied on the unstructured clinical notes found in electronic health records (EHR) as they easily summarize all the important observations about the patient state \cite{boag2018s}. However, clinical note in its raw unstructured form can lead to learning misleading representations due to noise present as, \textbf{1)} negated concepts \cite{miotto2016deep}, \textbf{2)} medical terms having different semantic roles \cite{kogan2005towards} in different context, and \textbf{3)} same medical term belonging to different contexts (e.g., history of present illness and family history).  

\begin{figure}[b!]
    \centering
    \includegraphics[width=0.40\linewidth]{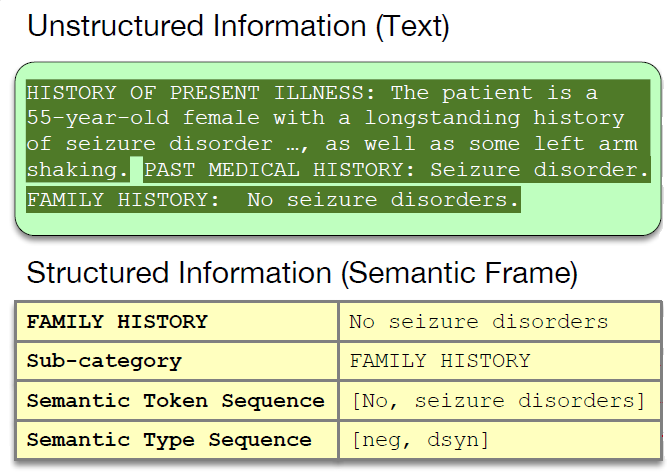}
    \caption{Sample clinical note (in green) and its corresponding semantic frame (in yellow). The semantic frame contains structured information extracted from the note and MetaMap. Here the semantic frame for only the last sentence in the note is depicted. Appendix contains the figure with all the semantic frames.}
    \label{fig::figure_2}
\end{figure}

Consider the clinical note shown in Figure \ref{fig::figure_2}. Like done in earlier work \cite{de2014medical}, simply extracting the structured sequence of UMLS concepts (e.g., seizure disorders....) using a biomedical concept identification system (e.g., MetaMap \cite{aronson2001effective}) to feed as input features into an embedding model fails to accurately reflect the exact visit information as it won't include the ``No" before the concept to assert the absence of the condition. This is because medical ontologies do not include mapping for many non-medical terms. Also, the fact that the medical condition ``seizure disorder" was present in the patient's present and past but not in family medical record will not be properly embodied by this embedding approach similarly due to missing context. On the other hand, learning directly from the sequence of texts with general-purpose embedding model (e.g., Word2Vec) \cite{minarro2014exploring}, although provides context, fails to capture the medically relevant semantic role of each term. Considering the merits and demerits offered by each approach independently, it calls for a joint approach to learn from both structured and unstructured information contained in clinical notes.


In Natural Language Understanding (NLU) works \cite{xu2013convolutional, zhang2018joint}, \textit{semantic frame} refers to the semantic  knowledge extracted from the text. As the medical concepts in the UMLS can be categorized to a semantic type and are linked with one another through semantic relationships, we call this information the semantic frame of the clinical text. \textit{Semantic Correspondence} is a commonly used technique in Computer Vision \cite{ zhang2016semantics, yang2017object} to semantically correlate the pixels in instances of an image that are similar at a high-level (e.g., same class) but vary in terms of appearance and geometry. In a similar vein, our goal in this paper is to adopt Semantic Correspondence for EHR texts in order to learn semantically-plausible representations capturing the correspondence between the text and its semantic frame, that enrich the representations with meaningful semantic associations, using patient mortality prediction as the motivating application. 

Moreover, as clinical notes are also documents, they have the hierarchical structure of documents (i.e., words form sentences, sentences form document), so we likewise inject this  knowledge of hierarchical structure in order to learn better note representation. In this direction, we propose a hierarchical embedding model (HierSemCor) that learns note representations through  correspondence of semantic understanding and contextual reasoning between texts and their semantic frames (i.e., note category/sub-category, UMLS semantic token/type sequences) in the clinical notes at multiple semantic levels. The proposed model HierSemCor, as shown in Figure \ref{fig::figure_1}, comprises three major components:

\begin{itemize}
\item{An unstructured embedding module that relies on Self-Similarity Matrix (SSM) \cite{ qiu2015convolutional} to capture the compositional and contextual information of clinical texts.}

\item{A structured embedding module that uses multi-layer ConvNet \cite{zhang2017deconvolutional} to obtain the semantic representation from different structured information associated with the text.}

\item{A hierarchical semantic correspondence learning module that first models the interaction between the text-semantic frame representations at the word-level to learn word embeddings, and then aggregates those to a sentence representation. The sentence representations are then enhanced with semantic knowledge from the structured information, which are similarly aggregated to a note representation.}
\end{itemize}

The contributions of this work are summarized as follows,
\begin{enumerate}
    \item We propose a novel model, HierSemCor designed for mortality prediction, that learns clinical note representations capturing its hierarchy and semantics simultaneously. To the best of our knowledge, this is the first work that employs hierarchical semantic correspondence on texts.  
    \item HierSemCor leverages the parallel structured information of the text in order to learn better representations. 
    \item Experimental analysis against multiple standard benchmarks for post-discharge patient mortality prediction tasks shows that our proposed framework performs better. 
\end{enumerate}

\section{The Proposed Model}
\begin{figure}[t!]
    \centering
    \includegraphics[width=0.7\linewidth]
    {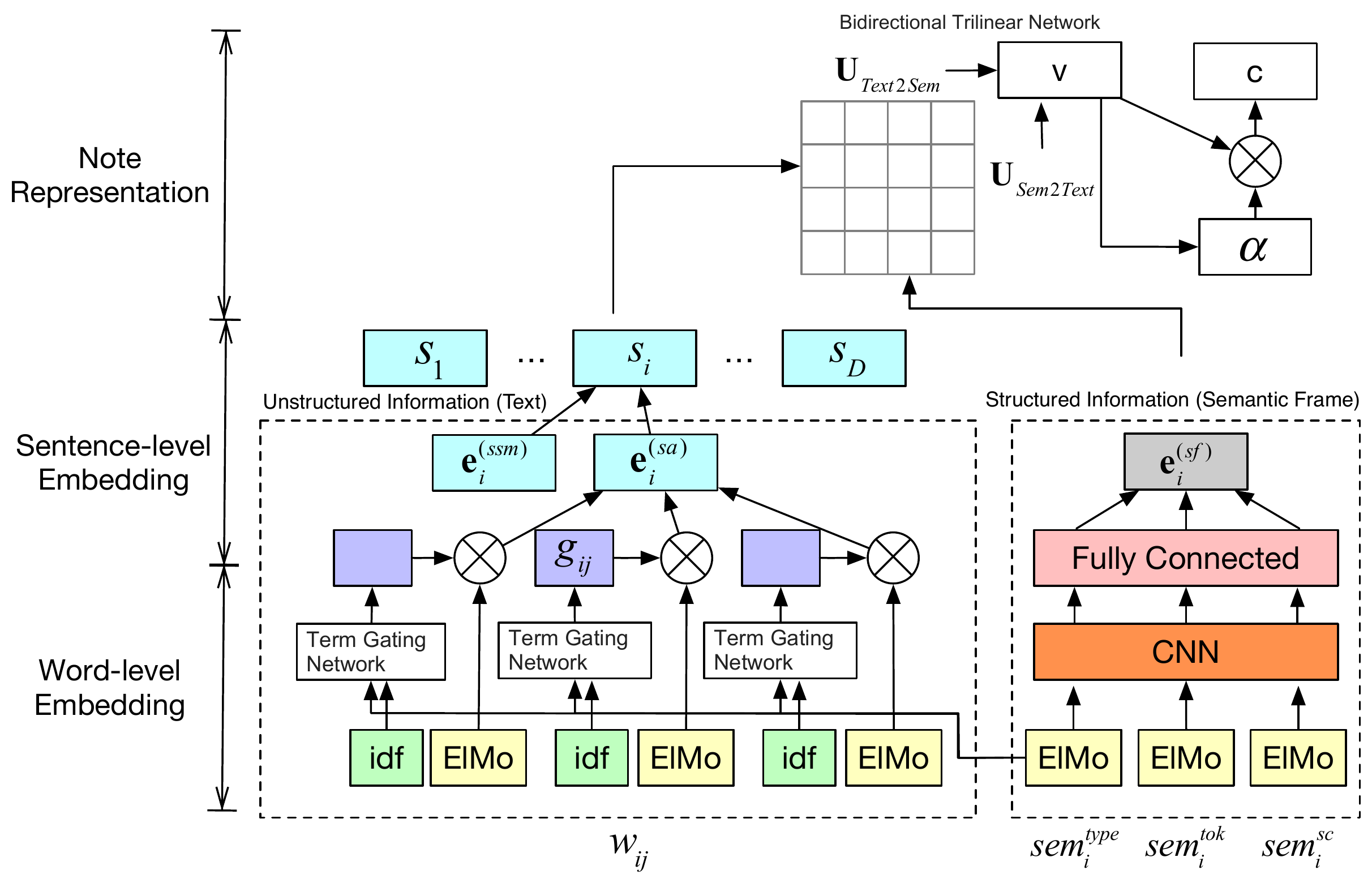}
    \caption{The proposed HierSemCor model.}
    \label{fig::figure_1}
\end{figure}

In this section, we introduce the proposed hierarchical semantic correspondence model, HierSemCor, which integrates text/semantic frame modeling and their semantic matching at multiple levels into a single model. The detailed architecture of HierSemCor is shown in Figure \ref{fig::figure_1}. The whole process of learning representations of notes can be decomposed into a hierarchy of three level embeddings. It first embeds each word by measuring its importance relative to the corresponding semantic frame token with a Term Gating Network (Sec 2.1). The aggregation of the importance-weighted word embeddings is then enriched with compositional patterns through a Self-Similarity Matrix-based module and its semantic frame equivalent similarly with a ConvNet-based Module, where the embedded sentence-semantic frame pair is then passed through a Bidirectional Trilinear Network to fuse their interactions into the sentence embedding (Sec 2.2). Finally, the sentence embeddings are summed, weighted by their salience score, to produce the semantic-aware note representation (Sec 2.3). As a result of the multi-level embeddings, both semantic matching signals and interactions between the unstructured text and its structured equivalent are incorporated into the final representation. The learning process is guided by a joint loss function, where the contrastive max-margin loss at the sentence level and smooth $L1$ loss at the note level are integrated (Sec 2.4).

Given a note $n = \{s_1...,s_i,...,s_D\}$ consisting of sentences $s_{i}$, where $D$ is the total number of sentences in the note and each sentence $s_i$ consists of words $\{w_{ij}|j = 1,2,...N\}$, the input into HierSemCor is a note, $n$, and its corresponding semantic frame, ${sem}= \{{sem}_{1},...,{sem}_{i},...,{sem}_{D}\}$. The different modules are described in the following subsections.

 \subsection{Term Gating Network} Textual noise in clinical notes can appear as unknown words, non-words and poor grammatical sentences. For example, unknown words could be complex medical terminologies, misspellings, acronyms and abbreviations, whereas non-words are clinical scores and measures \cite{nguyen2016text}. To reduce this noise, we weight the importance of each word with respect to its level of semantic interactions with the structured sequence tokens using a Term Gating Network \cite{ mcdonald2018deep}, as the parallel structured knowledge can provide cues towards the relevance of the word. 
 
 Given a sentence $s_i$ $(i = 1,2,...,D)$ consisting of words $\{w_{ij}|j = 1,2,...N\}$ and its corresponding semantic frame ${sem}_i$ having tokens $\{t_{ij}|j = 1,2,...N\}$, the gating mechanism to compute importance score of word $w_{ij}$ from the viewpoint of the structured information is defined as:
  \begin{equation}
 \large g_{ij} = \frac {exp(\mathbf{w}^T_g[t_{ij};idf({w_{ij}})])}{\sum\limits_{r=1}^{N}{exp(\mathbf{w}^T_g[t_{ir};idf(w_{ir})])}},
 \end{equation}
where $t_{ij}$ is the the $j$-th term of the semantic frame ${sem}_{i}$, $idf(w_{ij})$ is the inverse document frequency of the $j$-th word of sentence $s_i$, $\mathbf{w}_g$ is a weight vector and $;$ denotes concatenation. This aggregation weight will filter out textual noise by giving higher weight to more important words relative to the corresponding semantic frame. idf is included to remove commonly occurring words that only serve functional purpose. 

 The semantic-aware embedding of the sentence $s_i$ can then be obtained by summing up each ELMO \cite{peters2018deep} word embedding $\mathbf{e}(w_{ij})$ weighted by their semantic-aware importance weight:
 \begin{equation}
 \large \mathbf{e}^{(sa)}_i = \sum\limits_{j=1}^{N}{ g_{ij}\mathbf{e}(w_{ij})}.
 \end{equation}
 
\subsection{Sentence-level Embeddings}
\subsubsection{Structured Embedding Module}
The semantic frame for each sentence in the note can be decomposed into three parts --- ``note category and sub-category", ``UMLS semantic token sequence" and ``UMLS semantic type sequence", as illustrated in Figure \ref{fig::figure_2}. The ``note category" refers to the type of note (e.g., Discharge Summary, Nursing, Radiology) and ``note sub-category" is the section in the note out of the commonly found sections for that category that the sentence was mentioned under. Table 1 in Appendix shows the sections that can be found in Discharge Summary notes. This would explicitly disambiguate the status (e.g., present/absent) of the medical entity/condition as there is possibility of it being mentioned across different time events in the encounter (e.g., Family History, History of Present Illness). ``UMLS semantic token sequence" and ``UMLS semantic type sequence" describe the list of tokens in the sentence that are mapped to their UMLs semantic type by MetaMap, respectively. The semantic type for each token will provide cues for the specific semantic tag that the token is associated with in the UMLS.

Let ${sem}^z$, where $z \in \{sc, tok, type\}$,  denote the  ``note category and subcategory'', ``semantic token sequence'' and ``semantic type sequence'' respective components of the semantic frame. For simplicity of notations, we drop the $i$ subscript and refer to the semantic frame as $sem$.  We use multi-layer convolutional neural networks (ConvNets) \cite{zhang2017deconvolutional} to encode each component of the semantic frame as, $sem^{sc}$, $sem^{tok}$ and $sem^{type}$. Considering each component, ${sem}^z$, as a sequence of tokens, of length \textit{n} (zero padded), let $w^z_j$ denote the $j$-th token of ${sem}^z$. To learn a \textit{d} dimensional embedding vector for each token $w^z_j$ with a vocabulary size \textit{V}, embedding matrix $\mathbf{W}^{^z}_{emb} \in  \mathbb{R}^{d\times V}$ is trained jointly with the whole framework, where $\mathbf{W}^{^z}_{emb}[w^z_j]$ denotes the embedding vector of token $w^z_j$ for semantic component ${sem}^z$. Recent contextual word embedding model, ELMO \cite{peters2018deep}, trained on clinical notes and other documents in the clinical domain \cite{zhu2018clinical}, was used to initialize the weights of $\mathbf{W}^{^z}_{emb}$. The learned embedding vectors of all tokens of a sequence of length \textit{N} are then concatenated to represent it as $\mathbf{S}^{^z} \in \mathbb{R}^{d\times N}$. 

Our CNN architecture consists of total \textit{L} layers such that the \textit{L-1} are convolutional layers and the \textit{L-th} layer is a fully connected layer to encode the sentence into a fixed dimensional latent representation $h^z$. Each layer $l \in \{1,...,L\}$ uses $f_l$ filters, where the $i$-th filter has filter size $v_l$ and stride length $s_l$.  The first layer (\textit{l} = 1) applies convolution operation with a filter, $\mathbf{W}^{^z}_{conv}(i,l)  \in \mathbb{R}^{d\times {v_l}}$, on the sequence embedding $\mathbf{S}^{^z}$ to generate latent feature maps $m^{^z}(i, l) = f(\mathbf{S}^{^z}* \mathbf{W}^{^z}_{conv}(i,l)  + b^{^z}_{conv}(i,l))$ of dimensions $(n-v_l)/s_l + 1$. Here \textit{f} is a non-linear activation function, $b_{conv}(i,l) \in \mathbb{R}^{(n-v_l)/s_l + 1}$ are bias and * denotes convolution operation.  The feature maps for all the filters of layer $l$ are concatenated to obtain feature map $\mathbf{M}^{^z}_l = [m^{^z}(1,l),...,m^{^z}(f_l, l)] \in \mathbb{R}^{f_l\times [(n-v_l)/s_l + 1]}$. After the first layer, for each subsequent layer \textit{l}, the convolution operation is applied on feature map $\mathbf{M}^{^z}_{l-1}$ from the previous layer to yield feature map of that layer, $\mathbf{M}^{^z}_l$. We set the filter size of the $L-1$-th convolutional layer to $(n_l-2-v_l)/s_l + 1$, so that the spatial dimension in the $L$-th layer is collapsed to remove spatial dependency, and dimensions of the latent representation $\mathbf{h}^{^z}$ only depends on the number of filters in layer \textit{L}. 


The semantic frame representation, $\mathbf{e}^{(sf)} \in \mathbb{R}^{d_s}$, where $d_s$ = $3{f_L}$, is then computed as the concatenation (;) of the latent representations $\mathbf{h}^{^{sc}}, \mathbf{h}^{^{tok}}, \mathbf{h}^{^{type}} \in \mathbb{R}^{f_L}$:
\begin{equation}
 \large \mathbf{e}^{(sf)} = [{\mathbf{h}^{sc}}; {\mathbf{h}^{tok}}; {\mathbf{h}^{type}}].
 \end{equation}

\subsubsection{Unstructured Embedding Module}
 Patient information within a discharge summary note appear in distinct patterns. For example, general information of the patient (e.g., gender, birth date) appear as the first lines, followed by past and present status of the illness, concluding with the medication and discharge information as the last segments. Furthermore, each section is characterized by a common structure. Sections such as medications and vital signs are provided as a sequence of medications/vital signs and their values.  Whereas, other sections (e.g., History of Present Illness) are more descriptive.  In order to capture this underlying structure within each section of the clinical note, a Self-Similarity Matrix (\textbf{SSM}) is constructed. 
 
 For a note \textit{n} with $D$ sentences, its Self-Similarity Matrix is defined as ${\mathbf{SSM}}_{n} \in \mathbb{R}^{D\times D}$. Each element $({ssm_{n}})_{i{i'}} = {\mathbf{SSM}_{n}}(\mathbf{s}_i, \mathbf{s}_{i'})$, for ${i'} \in \{1,..,D\}$, is the cosine distance between the embedded sentences $\mathbf{s}_i$ and $\mathbf{s}_{i'}$. The sentence representations are obtained with pre-trained ELMO model. Each row of $\mathbf{SSM}_{n}$ exhibits the structural similarity of a text relative to other texts in the note. For a sentence $s_i$, given a window of size \textit{w}, a sub-matrix ${\mathbf{SSM}_{n}}(i-w:i+w,:)$ comprising entries for \textit{i} and \textit{w} lines above and below \textit{i} is extracted. This is to incorporate the structural nuances of the neighboring texts, alongside serving as context. We set \textit{w} = 1, and for the first and last lines we take $2w$ lines below and above it respectively so that all sub-matrices are of same dimensions. This sub-matrix is then passed through a convolution layer followed by a dropout layer and dense layer to get the corresponding representation $\mathbf{e}^{(ssm)} \in \mathbb{R}^{d_{ssm}}$. 

To get the final unstructured representation of the sentence, $\mathbf{e}^{(s)} \in \mathbb{R}^{d_s}$, $\mathbf{e}^{(ssm)}$ is concatenated to the semantic-aware importance weighted embedding of the sentence $\mathbf{e}^{(sa)}$:
\begin{equation}
 \large \mathbf{e}^{(s)} = [{\mathbf{e}^{(ssm)}};\mathbf{e}^{(sa)}].
 \end{equation}

 \subsubsection{Modeling Semantic Correspondence} Given the sentence embeddings $\mathbf{e}^{(s)}_{i}$ $(i = 1,2,...,D)$, they are passed through a Bidirectional Trilinear Network \cite{yu2018qanet} to fuse information between the sentence embeddings and their corresponding semantic frames. To facilitate this, it computes the trilinear features of the sentences to construct a similarity matrix, in order to assess the level of semantic interaction with their semantic frame, and vice-versa. It is composed of bidirectional attention in the form of sentence to semantic frame attention matrix, $\mathbf{U}_{Text2Sem} \in \mathbb{R}^{D\times h}$, and semantic frame to sentence attention matrix, $\mathbf{U}_{Sem2Text}\in \mathbb{R}^{D\times h}$. 
 
Let  ${\mathbf{e}_{k}^{(sf)}}$ $(k = 1,2,...,D)$ be the corresponding structured semantic frame embeddings of the unstructured sentences $s_i$ $(i = 1,2,...,D)$. A similarity matrix  ${\mathbf{SIM}} \in \mathbb{R}^{D\times D}$ is first constructed from the trilinear features defined as:
\begin{equation}
 \large {sim}_{ik} = \mathbf{W}_s[\mathbf{e}_{i}^{(s)};\mathbf{e}_{k}^{(sf)};\mathbf{e}_{i}^{(s)}\odot\mathbf{e}_{k}^{(sf)}],
 \end{equation}
 
 where ${sim}_{ik}$ is an element of the similarity matrix, $\mathbf{SIM}$, computed from the $i$-th sentence and $k$-th semantic frame, $\mathbf{W}_s$ is a trainable weight variable, $;$ denotes concatenation and $\odot{}$ denotes element-wise multiplication.

Each row and column of $\mathbf{SIM}$ are then normalized by applying softmax function to get normalized matrices $\bar{\mathbf{U}}_{Text2Sem} \in \mathbb{R}^{D\times D}$ and $\bar{\mathbf{U}}_{Sem2Text} \in \mathbb{R}^{D\times D}$ respectively:
\begin{equation}
 \bar{\mathbf{U}}_{Text2Sem} = softmax_{row}(\mathbf{SIM}),
  \end{equation}
  
 \begin{equation}
 \bar{\mathbf{U}}_{Sem2Text} = softmax_{column}(\mathbf{SIM}),
 \end{equation}
 
 where $softmax_{row}$ and $softmax_{column}$ are the respective row-wise and column-wise softmax operators. 

Using the normalized similarity matrices, the bidirectional attentions are computed as
 \begin{equation}
 {\mathbf{U}_{Text2Sem}} = \bar{\mathbf{U}}_{Text2Sem}\cdot {\mathbf{E}^{(sf)}}^T,
 \end{equation}
 
  \begin{equation}
 {\mathbf{U}_{Sem2Text}} = \bar{U}_{Text2Sem}\cdot\bar{U}_{Sem2Text}^T\cdot {\mathbf{E}^{(s)}}^T,
 \end{equation}
where $\mathbf{E}^{(s)}\in \mathbb{R}^{D\times {d_s}}$ is the sentence embedding matrix and $\mathbf{E}^{(sf)} \in \mathbb{R}^{D\times {d_s}}$ is the semantic frame embedding matrix. To get the final fused sentence representations $\mathbf{V}\in \mathbb{R}^{D\times 2h}$, we concatenate the Hadamard Products between $\mathbf{E}^{(s)}$ and its corresponding attention matrix $\mathbf{U}_{Text2Sem}$, and $\mathbf{E}^{(sf)}$ and its attention matrix $\mathbf{U}_{Sem2Text}$:
 \begin{equation}
 {\mathbf{V}} = [\mathbf{E}^{(s)}\odot \mathbf{U}_{Text2Sem};\mathbf{E}^{(sf)}\odot \mathbf{U}_{Sem2Text}].
 \end{equation}
 
 \subsection{\textbf{Note Representation}}
 Clinical notes can have a lot of information in them and all of them might not be salient. Therefore, we compute a weighted representation for the note, $\mathbf{c}\in \mathbb{R}^{2h}$, from the learned sentence representations, where the weight indicates the level of salience the sentence contributes and is computed using a feed-forward neural network:
 \begin{equation}
 {\boldsymbol \alpha} = {\mathbf{W}^T_n}V + \mathbf{b}_n,
 \end{equation}
 
 \begin{equation}
 \mathbf{c} = \sum\limits_{i=1}^{D}{\alpha_i \mathbf{v}_i}.
 \end{equation}
 
\subsection{Joint Training} In the HierSemCor model, all the modules/networks
are jointly trained in an end-to-end fashion. For this purpose, we define the joint loss function in the training process upon the
contrastive max-margin loss to optimize at the sentence-level and the smooth $L1$ loss to optimize at the note-level:
\begin{equation}
\mathcal{L} = \mathcal{L}_{cmm} + \mathcal{L}_{{smooth}_{L1}}.
\end{equation}

 Contrastive Max-margin objective function \cite{socher2013reasoning} works by assigning a higher score to true sentence-semantic frame pairs than their negative sample pairs. Negative sample pairs for each sentence-semantic frame pair in the training set is formed by replacing the semantic frame of the sentence with a random semantic frame of same length as the sentence. It is defined as:

\begin{equation}
\mathcal{L}_{cmm} = \sum\nolimits_{(s,sem) \in C_i}^{} {\sum\nolimits_{(s,sem') \in {C'_i}}^{}} {[1 - score(s,sem) + score(s,sem')] + \lambda ||\theta ||_2^2}.
\end{equation}

Here $score$ is the semantic similarity score and is considered as the corresponding salience score $\alpha{}_i$ for the sentence $s_i$. So $score(s_i,{sem}_i)$ is the $\alpha{}_i$ score for the true sentence-semantic frame pair from the training set, \textit{C}, containing all true sentence-semantic frame pairs. $score(s_i,{sem}_i')$, on the other hand, is that of a negative sample from the set of sentence-corrupted semantic frame pairs $\textit{C'}$. $\theta$ denotes all the parameters of the sentence-embedding module. Standard $\ell{2}$ regularization of the parameters is used, weighted by the hyperparameter $\lambda$.

The smooth $L1$ loss is computed between the generated note representation, $\mathbf{c}$, and the flattened ELMO representation, $\mathbf{s}'$, of the sentence embedding matrix and is defined as:


 \begin{equation}
 \mathcal{L}_{{smooth}_{L1}}(\mathbf{c}-\mathbf{s}') = 
 \begin{cases}
    0.5\|\mathbf{c}-\mathbf{s}'\|_2^2,& \text{if } \|\mathbf{c}-\mathbf{s}'\|\leq 1\\
    \|\mathbf{c}-\mathbf{s}'\|_1 - 0.5,& \text{if } \|\mathbf{c}-\mathbf{s}'\|\geq 1
\end{cases}
\end{equation}

 \section{Patient Mortality Prediction}
We frame the problem of predicting whether the patient will die after hospital discharge as a binary classification task. Assuming that all the representations, $<\mathbf{c}_{1}^p,...,\mathbf{c}_{|p|}^p>$, of the notes associated with a patient $p$ are aggregated to a final patient representation, $\mathbf{pat}_p$,  the patient mortality prediction task tries to learn the following mapping function,
\begin{equation}
 \phi: \mathbf{pat}_p \rightarrow{} y_p,
\end{equation}
where $y_p \in \mathbb{Y} = \{alive, dead\}$. The prediction is done in two scenarios, the patient will die within (1) 30 days and (2) 1 year after discharge from the hospital. 

 \section{Experiments}
  \subsection{Data}
  The MIMIC-III dataset \cite{johnson2016mimic} contains de-identified clinical records for $>$ 40K patients admitted to critical care units over a period of 11 years. It includes around 2,000,000 clinical notes, with multiple notes associated with each patient. Notes can be recorded for different purpose (e.g., Physician, Nursing, Echo, ECG, Radiology,
Nutrition, Pharmacy, Discharge Summary, etc.). We restrict to adults ($\geq$ 18 years old) with only one hospital admission in order to avoid any explicit references to label information when they depend on the discharge time. We consider notes of the patient's first hospital stay from the following categories: Nursing, Discharge Summary, Radiology, Echo, ECG and Physician. The Social Security Master Death Index in MIMIC III was used to find the exact post-discharge mortality date and create the positive class instances. As there is a high imbalance in the dataset due to fewer positive samples (i.e., dead), we subsample from the negative instances (i.e., alive) to reduce skewness. Table \ref{tab::Table1} outlines the statistics about the dataset. We divided this dataset into training, test and validation sets randomly with split of 80\%-10\%-10\%.  

\begin{table}[hbtp!]
\centering
\caption{Data Statistics Summary of Patient Mortality Post Hospital Discharge}
\label{tab::Table1}
\resizebox{0.5\linewidth}{!}{
\begin{tabular}{llll}
\toprule
\textbf{MIMIC-III} & \textbf{NUMBER}\\ \midrule
\# of total patients    & 9924\\
\# of total notes                 & 20244  \\
avg. \# of notes per patient               & 2.2  \\ 
\# of patients who died within 30 days & 1156      \\
\# of patients who did not die within 30 days & 2500   \\
\# of patients who died within 1 year &  3768  \\
\# of patients who did not die within 1 year & 5000   \\
\bottomrule
\end{tabular}
}
\end{table}

 \subsection{Baselines}
 The proposed HierSemCor model outperforms all baselines and we validate it through experimental comparison with both state-of-the-art feature-based and deep learning models in three types of EHR data settings: the model is trained on 1) only structured data, 2) only unstructured data and 3) a combination of both structured and unstructured data. All the models are evaluated using the area under the ROC curve (AUC-ROC) for the patient mortality task in the two scenarios. The ROC curve gives us insight into the trade-off between the true positive rate and the false positive rate at different thresholds for different models.

\begin{itemize}
  \item[] \textbf{BoCUI}: This model \cite{sushil2018patient} represents each note as a bag-of-medical-concepts. The corresponding tf-idf scores are used as the feature values. 
  \item[] \textbf{SDAE-BoCUI}: This model \cite{sushil2018patient} uses a stacked denoising autoencoder with the bag-of-medical-concept representation of the note as the input. 
  \item[] \textbf{Concept CNN}: It is a convolutional neural network based model. The architecture comprises three convolutional layers and three max-pooling layers. Medical concepts in the UMLS extracted from the note are passed as input.
  \item[] \textbf{LDA based Concept SVM}: It first trains an LDA \cite{blei2003latent} with 100 topics to get the topic representations of the note. Then  SVM \cite{hearst1998support} is trained on the topic distribution features . 
  \item[] \textbf{BoW}: This model \cite{sushil2018patient} represents each note as bag-of-words.
  \item[] \textbf{SDAE-BoW}: This model \cite{sushil2018patient} uses a stacked denoising autoencoder with the bag-of-words representation of the note as the input.
  \item[] \textbf{Doc2Vec}: \cite{le2014distributed} learns distributed representation of the note using an unsupervised model.
  \item[] \textbf{LDA based Word SVM}: Same as \textbf{LDA based Concept SVM}, but generates concept-level topic distributions of the document. 
  \item[] \textbf{Text CNN}: It is a convolutional neural network model based on the model proposed in, with the architecture same as Concept CNN. 
  \item[] \textbf{Combined LDA with SVM}: The method is same as \textbf{LDA based Word SVM}, but uses both concepts and notes to get the topic distribution from LDA. 
  \item[] \textbf{BK-DNN}: The Basic Knowledge-aware Deep Dual Network model \cite{liu2019knowledge} consists of two CNN-based sub-neural networks - one to learn the text representation and the other to learn the concept representation of the note. The two representations are then fused together. It does not take into consideration the structural and compositional regularities present in clinical notes as is addressed in the proposed HierSemCor.  
  \item[] \textbf{AK-DNN}: Built on top \textbf{BK-DNN} model with a co-attention mechanism.
\end{itemize}

 \subsection{Results and Discussion}
Table \ref{tab::Table3} presents the overall performance of all methods on the patient post-discharge 30-day and 1-year mortality prediction tasks. One general observation that holds true for all the methods is that prediction further in the future is more difficult, as is indicated by the lower scores for the 1-year mortality in comparison to 30-day mortality task. BoCUI and BoW, which are sparse representations of the clinical note, can be seen to give sub-par performance relatively as they discard any information describing the structure and contexts between sentences in the note that can provide helpful semantic features for the classification. Passing these sparse representations through non-linearity, however, shows improvement with stacked denoising autoencoder variants (i.e., SDAE-BoCUI, SDAE-Bow). The structured counterparts of the bag-of-words variants (i.e., BoCUI, SDAE-BoCUI) are not able to generalize as well due to missing mapping of important unstructured tokens to equivalent UMLS concepts. Using CNN to encode the note shows its effectiveness empirically for Concept CNN and Text CNN models and justifies our use of ConvNet to encode the note and its semantic frame. The low AUC-ROC results of the LDA based models (i.e., LDA based Concept SVM, LDA based Word SVM) suggest that representing the document in terms of its topic distributions does not suffice performance gain as it is unable to capture the comprehensive lexical knowledge present among the sentences. Powerful document embedding model, Doc2Vec, although captures contexts, but misses the syntactic nuances of sentences specific to the structure of clinical notes. State-of-the-art models BK-DNN and AK-DNN also supplement the unstructured textual representation of note with its structured medical concepts, however, HierSemCor differs from it in the following ways, (1) HierSemCor explicitly incorporates the inherent hierarchical structure information of the note through learning embeddings at different levels, (2) structured information is provided not just as the corresponding UMLS medical concepts, but also the semantic type of the medical concepts to better reflect the structured semantics and (3) we include both relevance and semantic matching signals with the Term Gating and Bidirectional Trilinear Networks respectively, that better captures the interactions and strengthens the note representations. We can say that these factors lead to HierSemCor outperforming these state-of-the-arts.  

 \newcommand{\STAB}[1]{\begin{tabular}{@{}c@{}}#1\end{tabular}}
 
\begin{table}[htb!]
\centering
\caption{Performance of different models on Patient Mortality Prediction Task within 30 days and 1 year after discharge from hospital. We have compared the proposed HierSemCor model against baselines which only use the structured data (Structured), only use the unstructured data (Unstructured) and both structured and unstructured data (Combined).}
\label{tab::Table3}
\begin{tabular}{l|ll}
\toprule
\textbf{MODEL}        & \textbf{30 DAYS}       & \textbf{1 YEAR}                \\ \midrule
\textbf{Structured}  & 	&\\
\phantom{00}BoCUI  & 0.679	& 	0.501         \\
\phantom{00}SDAE-BoCUI  &  0.788	&	0.772   \\
\phantom{00}Concept CNN  & 0.802	& 	 0.775 \\
\phantom{00}LDA based Concept SVM  & 0.698	& 0.695	  \\
\hline
\textbf{Unstructured}  & 	& \\
\phantom{00}BoW  &  0.784		&  0.562  \\
\phantom{00}SDAE-BoW  & 0.811	& 	0.789  \\
\phantom{00}Text CNN  & 0.832	& 0.805	  \\
\phantom{00}LDA based Word SVM  & 0.755	& 0.723 \\
\phantom{00}Doc2Vec  & 0.813	& 	 0.769 \\ \hline

\textbf{Combined}  & 	& \\
\phantom{00}LDA with SVM  & 0.774 & 0.748 	         \\ 
\phantom{00}BK-DNN  & 0.844	& 0.830	 \\ 
\phantom{00}AK-DNN  & 0.865 & 0.841   \\ \hline
HierSemCor$_{\text{w/o. Unstruct}}$  & 0.840 & 0.816  \\ 
HierSemCor$_{\text{w/o. Struct}}$  & 0.857 & 0.834  \\ 
HierSemCor  & \textbf{0.881} & \textbf{0.852}  \\ 
\bottomrule
\end{tabular}
\end{table}

\begin{figure*}[bt!]
    \centering
    \includegraphics[width=0.8\linewidth]{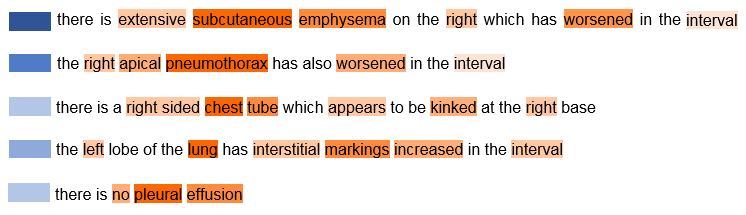}
    \caption{Visualization of the importance weights and salience weights of words and sentences in a clinical note, respectively, learned by HierSemCor. }
    \label{fig::figure_3}
\end{figure*}
 \subsection{Ablation Study} To further demonstrate the contribution of modeling the semantic correspondence between the text-semantic frame, we perform experiments with ablated versions of HierSemCor, as depicted in the bottom part of Table \ref{tab::Table3}. $HierSemCor_{{w/o}_{.Struct}}$ is our model without the Structured Embedding Model, that just takes the unstructured text as input and passes the output of the Unstructured Embedding model through a softmax layer; $HierSemCor_{{w/o}_{.Unstruct}}$ is our model without the Unstructured Embedding Model, that just
takes the semantic frame embedding as input and passes the output of the Structured Embedding model through a softmax layer; 
It can be seen that the proposed model has significant AUC-ROC performance gain over both the ablated models. This indicates that leveraging the interactions between the structured and unstructured information is indeed more beneficial than their respective
independent representation.
 
 \subsection{Qualitative Analysis} We show through visualization of the importance weights, $g_{it}$ and the degree of informativeness of the sentence, $\alpha_i$, how HierSemCor learns semantic representation of the note. Figure \ref{fig::figure_3} depicts a clinical note of a patient who has the gold-truth label of ``dead" for the 30-days mortality prediction task. Every individual line stands for a sentence in the note. The importance weight of a word is denoted in orange and the salience weight of the sentence is denoted in blue. The hue of the color indicates the value of the weight, with darker meaning higher value. As the word importance weight is computed relative to the semantic frame, we can see that words with available semantic mapping in UMLS have been assigned the highest weights. For example, medical terminologies like ``subcutaneous", ``emphysema" and ``pneumothorax". Words describing the status of the medical condition (e.g., ``worsened", ``increased"), indicate deterioration and the possibility of mortality, are also captured. As ``no" is an important contextual word but does not have corresponding structured mapping in UMLS, we feed a special "neg" tag as its semantic type. HierSemCor is therefore able to focus on such words as well.      
 
 \section{Related Works}

\textbf{Semantic Correspondence} In the Computer Vision domain, establishing semantic correspondence between pixels in different instances of object/scene images is a seminal task with wide applications. In \cite{yang2017object} a pixel-to-pixel dense correspondence model, Object-aware Hierarchical Graph (OHG), was proposed to estimate correspondence from semantic to low-level with a discriminative classifier assigned to each grid cell and a guidance layer. \cite{han2017scnet} used a CNN-based model to learn semantic correspondence with region proposals as the matching primitives. Whereas in \cite{bristow2015dense}, instead of using similarity kernels to compute semantic correspondence, LDA classifier was used to measure the likelihood of two points matching. 

Use of semantic correspondence for text analysis is at a nascent stage. In \cite{liang2009learning}, a generative model was used to learn the semantic correspondence between a text and world state represented as a set of records. \cite{jung2018learning} learned semantic representations for NLU based on correspondence and reconstruction losses with LSTM encoder. HierSemCor, however, applies semantic correspondence at multiple levels to capture the interactions among the different textual entities (word-word, sentence-sentence).  

\textbf{Patient Mortality Prediction} There are works which have used either the structured or the unstructured information, or both from the electronic health records for this problem. \cite{zhang2014data} used the structured data consisting of static and time course measurements (i.e., demographic, injury, background information) to predict mortality for trauma patients using logistic regression. An ensemble machine learning technique was used in \cite{pirracchio2015mortality} with subset of numerical scores as the features to predict mortality of ICU patients. A two layer CNN trained on only unstructured medical notes was used in \cite{grnarova2016neural}. They additionally replicated the loss at intermediate steps to account for the long-term dependencies in notes. \cite{ghassemi2014unfolding} first used an LDA to generate the topic distribution of notes and then combined them with structured features, which are then passed through SVM for the prediction. A Labeled-LDA augmented with domain knowledge to guide the topic model learning was used in \cite{luo2016interpretable}. 

\section{Conclusion}
 In this paper, we propose HierSemCor as a novel hierarchical clinical note representation approach to model patient mortality prediction. HierSemCor explores a new direction to learn clinical note representation by applying semantic correspondence between the structured and unstructured entities at multiple levels. This leads to learning a more meaningful representation of the note, that is able to capture its semantic and compositional subtleties. Quantitative and qualitative evaluations of our model on two post-discharge patient mortality tasks demonstrate its effectiveness and robustness.

\bibliography{thebibliography}
\bibliographystyle{abbrv}
\end{document}